\documentclass[11pt, a4paper, logo, copyright, nonumbering]{ubi}

\usepackage{natbib}

\usepackage{CJKutf8}

\usepackage{xargs}  

\usepackage{todonotes}  

\usepackage{multirow}

\usepackage{cleveref}

\usepackage{amsmath}
\usepackage{dsfont}

\usepackage{subcaption}
\usepackage{wrapfig}

\usepackage{svg}
\usepackage[breakable]{tcolorbox}
\usepackage{fontawesome}
\usepackage{xcolor}

\usepackage[utf8]{inputenc}
\usepackage{booktabs}
\usepackage{graphicx}
\usepackage{array}
\usepackage{tabularx}
\usepackage{xcolor,colortbl}  
\definecolor{boxcolor}{HTML}{d92523} 
\definecolor{bulbcolor}{HTML}{e3b87f} 

\hypersetup{
    colorlinks=true,            
    linkcolor=blue,             
    filecolor=magenta,          
    urlcolor=cyan,              
    citecolor=purple,             
    pdftitle={Hyperball May Not Be a Free Lunch},
    pdfauthor={Yihao Xiao, Jialong Sun, Zitian Gao, Zeming Wei, Chutian Wang, Ran Tao, Jiaye Teng, Bryan Dai},
    pdfpagemode=FullScreen,
    }
 
\renewcommand{\today}{}
\newcommandx{\info}[2][1=]{\todo[linecolor=red,backgroundcolor=red!25,bordercolor=red,#1]{#2}}

\title{\centering Hyperball May Not Be a Free Lunch}

\author{
\normalfont Yihao Xiao\textsuperscript{1,5,\ensuremath{\dagger}},
Jialong Sun\textsuperscript{4,\ensuremath{\dagger}},
Zitian Gao\textsuperscript{1}, Zeming Wei\textsuperscript{2},
Chutian Wang\textsuperscript{3}, \\ \normalfont Ran Tao\textsuperscript{1},
Jiaye Teng\textsuperscript{5}, Bryan Dai\textsuperscript{1,*}\\ \vspace{4mm}
\normalfont\small
\textsuperscript{1}IQuest Research; \textsuperscript{2}Peking University;\\
\normalfont\small\textsuperscript{3}Sun Yat-sen University;\\
\normalfont\small\textsuperscript{4}Shenzhen University of Advanced Technology;\\
\normalfont\small\textsuperscript{5}Shanghai University of Finance and Economics\\
\normalfont\footnotesize\textsuperscript{\ensuremath{\dagger}}Equal contribution;
\textsuperscript{*}Corresponding author
}

\begin{abstract}
For scale-invariant deep networks, Hyperball-style optimizers have shown strong performance in large-scale training by fixing the norms of matrix-valued parameters and normalizing updates. However, the source of their advantage remains unclear. Starting from the angular displacement between consecutive parameter states, we derive an angular effective learning rate that accounts for the parameter-update angle, parameter norm, and update norm. We also show that the conventional norm-based measure is a special case under parameter-update orthogonality. We then decompose optimizer updates into radial and tangential components and analyze how radial updates affect one-step angular displacement. Under the training configurations considered, numerical results show that the radial component has only a limited direct effect on the angular effective learning rate. It therefore cannot explain why MuonH converges more slowly than MuonWD early in training but overtakes it later.
To further isolate the underlying mechanism, we devise a heuristic experiment that modifies only the learning-rate schedule so that the dynamics of each optimizer reproduce those of the other. The results suggest that their main difference stems from the evolution of the effective step size rather than an intrinsically superior update direction induced by Hyperball. Our pretraining experiments further show that more aggressive learning-rate decay can accelerate MuonH early in training but may impair its later performance. Thus, maintaining a constant angular velocity does not eliminate the learning-rate-scheduling problem; careful scheduling remains essential to realizing the potential of Hyperball-style optimizers. Our code is publicly available at \url{https://github.com/mangocrazz/hyperball-may-not-be-a-free-lunch}.
\end{abstract}

\begin{document}
\begin{CJK*}{UTF8}{gbsn}

\maketitle

\section{Introduction}
In deep networks with scale-invariant structure, parameter norms, learning rates, and optimizer update directions jointly determine how the model actually moves through parameter space~\citep{ballsun,1,2}. Recent optimizers with matrix constraints, such as MuonH~\citep{li2025normuon,cesista2025stiefel,rmnp,MuonH}, restrict matrix-valued parameters to norm spheres with prescribed radii in an effort to reduce the influence of parameter-scale variation on training dynamics. Although these methods have shown promising empirical performance in large-scale pretraining, the mechanism underlying their advantage remains unclear. In particular, it is not yet known whether the improvement arises from a more favorable update direction, from suppressing radial updates, or from a constraint-induced change in step size.

A common interpretation is that weight decay or norm constraints primarily control the effective learning rate~\citep{eff1,eff2,eff3}: as a parameter norm grows, scale invariance reduces the gradient magnitude relative to the parameter magnitude, thereby decreasing the effective update size. This interpretation, however, typically treats the effective learning rate mainly as a function of the parameter norm and implicitly assumes that the parameter and the optimizer update are approximately orthogonal. That assumption need not hold for modern optimizers that use momentum, preconditioning, or matrix-valued update rules. In such cases, the angle between the parameter and the update, the norm of the update, and the evolution of both quantities can all affect the actual angular displacement. Controlling the parameter norm alone therefore does not identify how Hyperball changes the optimization trajectory.

We study a representative empirical phenomenon in Hyperball training. Relative to its non-Hyperball counterpart, a Hyperball optimizer can exhibit distinctly phase-dependent behavior: it underperforms early in training but overtakes the baseline in the middle or later stages. This behavior is not fully explained by the simple claim that a fixed norm yields a stable effective learning rate. If Hyperball acts mainly by suppressing radial updates, why should that suppression produce a phase transition rather than a monotone difference in training? Conversely, if its principal effect is to alter the effective step size, can the same dynamics be reproduced solely by changing the learning-rate schedule? These questions are the central focus of this work.

To answer these questions, we characterize effective learning rates through the angular displacement induced by each parameter update. Specifically, we incorporate the angle between the parameter and the update into the analysis, derive an angular effective learning rate for a general update rule, and recover the familiar norm-based expression as a special-case approximation. We then separately examine the direct contribution of radial updates to the angular effective learning rate and the joint effect of Hyperball's constraints on update scale and parameter norm. Finally, we construct a state-dependent learning-rate matching procedure that aligns the angular effective learning rate of a non-Hyperball optimizer with that of its Hyperball counterpart at every step. This procedure tests the extent to which the training dynamics of the two optimizer families can reproduce one another. Overall, our contributions are summarized as follows:
\begin{itemize}
     
 \item \textbf{Conceptually}, we derive an angular effective learning rate that explicitly accounts for the angle between the parameter and the update, and we state the conditions under which the conventional effective-learning-rate approximation is valid.

\item \textbf{Theoretically}, we examine the role of radial updates through controlled numerical analyses. Under the training configurations considered in this work, their direct effect on the angular effective learning rate is relatively small.

\item  \textbf{Practically}, we propose a state-dependent learning-rate alignment method for isolating the dynamical difference between Hyperball and non-Hyperball optimizers. Our experiments show that part of Hyperball's phase-dependent training behavior can be reproduced by aligning effective learning rates.
\end{itemize}
\section{Preliminaries: Effective Learning Rate and Angular Updates}

This section introduces the update rule, notation, and geometric relations used in the subsequent analysis. Rather than characterizing an optimizer's effective step solely through the parameter norm, we explicitly account for the relative direction of the parameter and the update. This perspective shows that, once the update is not orthogonal to the parameter, the angular step depends not only on the learning rate and parameter norm but also on the update norm and the angle between the two.

\subsection{Generic update rule and notation}

Consider an arbitrary matrix-valued parameter block $W_t$, with each block analyzed independently. Let $U_t$ denote the update produced by the optimizer at step $t$. This quantity includes momentum, preconditioning, orthogonalization, and any other transformation internal to the optimizer, but excludes the external learning-rate multiplier. A generic update with decoupled weight decay can be written as
\begin{equation}
    W_{t+1} = \alpha_t W_t - \eta_t U_t,
    \qquad
    \alpha_t = 1-\eta_t\lambda_t,
    \label{eq:generic-update}
\end{equation}
where $\eta_t>0$ is the learning rate and $\lambda_t\geq 0$ is the weight-decay coefficient. For convenience, define the cosine similarity between $W_t$ and $U_t$ as
\begin{equation}
    c_t = \frac{\langle W_t,U_t\rangle_F}{\|W_t\|_F\|U_t\|_F}.
\end{equation}

\subsection{Scale invariance and the limitation of norm-only analysis}

For a scale-invariant parameter, as induced for example by a normalization layer, the loss satisfies~\citep{ballsun}
\begin{equation}
    \mathcal{L}(\rho W)=\mathcal{L}(W),
    \qquad \forall \rho>0.
    \label{eq:scale-invariance}
\end{equation}
Differentiating Eq.~\eqref{eq:scale-invariance} with respect to $\rho$ gives
\begin{equation}
    \left\langle
    W,\nabla_W\mathcal{L}(W)
    \right\rangle_F=0,
    \qquad
    \nabla_W\mathcal{L}(\rho W)
    =
    \frac{1}{\rho}\nabla_W\mathcal{L}(W).
    \label{eq:scale-invariant-gradient}
\end{equation}
Thus, if $U_t$ is the instantaneous gradient, then $c_t=0$. In this special case, increasing the parameter norm decreases the relative update size, which is the basis of the classical effective-learning-rate analysis.

For a modern optimizer, however, $U_t$ is generally not the instantaneous gradient. Momentum aggregates past gradients, preconditioning changes their relative scale across directions, and matrix optimizers may apply additional transformations to the update. Consequently, even when the raw gradient satisfies Eq.~\eqref{eq:scale-invariant-gradient}, it does not follow in general that $\langle W_t,U_t\rangle_F=0$. Describing the effective step solely by $\eta_t/\|W_t\|_F$ may therefore omit effects due to both the direction and the magnitude of the optimizer update.

\subsection{Angular effective learning rate}

We measure the directional displacement of an update by the angle between two consecutive parameter states:
\begin{equation}
    \Delta\phi_t
    =
    \angle(W_t,W_{t+1}).
    \label{eq:angular-change}
\end{equation}
This angular displacement excludes pure radial rescaling and is therefore particularly appropriate for analyzing optimization dynamics in scale-invariant networks. The following proposition gives the angular displacement under the generic update rule.

\paragraph{Proposition 1.}
For the update in Eq.~\eqref{eq:generic-update}, if
$\alpha_t\|W_t\|_F-\eta_t c_t\|U_t\|_F>0$, then
\begin{equation}
    \tan(\Delta\phi_t)
    =
    \frac{
        \eta_t \|U_t\|_F\sqrt{1-c_t^2}
    }{
        \alpha_t\|W_t\|_F-\eta_t c_t\|U_t\|_F
    }.
    \label{eq:exact-angular-update}
\end{equation}

The result follows directly by decomposing $U_t$ into components parallel and orthogonal to $W_t$. The numerator is the tangential component that rotates the parameter direction, whereas the denominator combines the radial component of the current parameter with the radial component of the update. When the single-step angular displacement is small, $\tan(\Delta\phi_t)\approx\Delta\phi_t$. We therefore define the angular effective learning rate as
\begin{equation}
    \eta^{\phi}_{\mathrm{eff},t}
    \triangleq
    \frac{\Delta\phi_t}{\|U_t\|_F}
    \approx
    \frac{
        \eta_t\sqrt{1-c_t^2}
    }{
        \alpha_t \|W_t\|_F-\eta_tc_t \|U_t\|_F 
    }.
    \label{eq:angular-effective-lr}
\end{equation}

Equation~\eqref{eq:angular-effective-lr} shows that the angular step is jointly determined by four time-varying quantities: the learning rate $\eta_t$, the parameter norm $\|W_t\|_F$, the update norm $\|U_t\|_F$, and the cosine similarity $c_t$. When $c_t=0$, Eq.~\eqref{eq:angular-effective-lr} reduces to
\begin{equation}
    \eta^{\phi}_{\mathrm{eff},t}
    \approx
    \frac{\eta_t}{\alpha_t \|W_t\|_F},
    \label{eq:classical-effective-lr}
\end{equation}
Equation~\eqref{eq:classical-effective-lr} is the conventional norm-based expression for the effective learning rate. It relies, however, on the assumption $c_t=0$, which need not hold for a transformed optimizer update. Section~\ref{sec:radial-update} examines how violations of this assumption affect the angular effective learning rate.

\subsection{Hyperball update}

Hyperball constrains both the parameter norm and the update norm. Let the radius of the constraint sphere be the initial norm $R=\|W_0\|_F$, and define the normalized update
\begin{equation}
    V_t = R\frac{U_t}{\|U_t\|_F}.
\end{equation}
The Hyperball update is then
\begin{equation}
    W_{t+1}
    =
    R\cdot
    \frac{W_t-\eta_t V_t}
    {\|W_t-\eta_t V_t\|_F}.
    \label{eq:hyperball-update}
\end{equation}
Equation~\eqref{eq:hyperball-update} ensures that $\|W_{t+1}\|_F=R$ after every step while also fixing the update norm at $\|V_t\|_F=R$. The update can equivalently be written as
\begin{equation}
    W_{t+1}
    =
    a_t W_t-\eta_t a_tV_t,
    \qquad
    a_t
    =
    \frac{R}{\|W_t-\eta_t V_t\|_F}.
    \label{eq:hyperball-decomposition}
\end{equation}

Equation~\eqref{eq:hyperball-decomposition} exposes two coupled operations in Hyperball: the optimizer update is first rescaled to a fixed norm, and the resulting parameter is then projected back onto the sphere of radius $R$ through the state-dependent coefficient $a_t$. Equivalently, $a_t$ induces both a time-varying radial rescaling of the parameter and a rescaling of the update. This coupling is the principal structural difference between Hyperball-style and conventional optimizers. Applying the same angular decomposition as above yields the following small-angle effective learning rate:
\begin{equation}
    \eta_{\text{eff},t}^{\text{H},\phi}\approx\frac{\eta_t\sqrt{1-c^2_t}}{(1-\eta_tc_t)R}
\end{equation}
The angular effective learning rate of a conventional optimizer depends jointly on $\eta_t$, $\|W_t\|_F$, $\|U_t\|_F$, and $c_t$. By fixing the parameter norm and normalizing the update, Hyperball removes the independent dependence on $\|W_t\|_F$ and $\|U_t\|_F$, leaving only $\eta_t$ and $c_t$. Under a prescribed learning-rate schedule, the remaining state-dependent variation is therefore carried by $c_t$. Section~\ref{sec:radial-update} uses this unified expression to test whether the radial component represented by $c_t$ materially changes the effective learning rate.

\section{Do Radial Updates Provide an Additional Benefit?}
\label{sec:radial-update}
The preceding analysis shows that the angular effective learning rate depends not only on the learning rate and parameter norm but also on $c_t$ and $\|U_t\|_F$. We now ask whether radial updates are the primary cause of MuonH's phase-dependent training dynamics. One intuitive explanation is that, when $c_t<0$, the optimizer update increases the parameter norm; by projecting the parameter back onto a sphere of fixed radius, Hyperball suppresses this radial motion and may thereby change the optimization speed in the early and middle stages of training. This explanation conflates two distinct mechanisms: the direct effect of a radial component on the current angular step and its indirect effect on future steps through changes in the parameter state. We separate these mechanisms and test whether the direct effect alone can explain the observed dynamics.

\subsection{Radial and tangential components}

Under Eq.~\eqref{eq:generic-update}, $U_t$ can be decomposed into a radial component parallel to the current parameter and a tangential component orthogonal to it:
\begin{equation}
\begin{aligned}
    &U_t^{\parallel}
    =
    \frac{\langle W_t,U_t\rangle_F}{\|W_t\|_F^2}W_t
    =
    \frac{c_t\|U_t\|_F}{\|W_t\|_F}W_t,
    &U_t^{\perp}
    =
    U_t-U_t^{\parallel}.\\
    &\|U_t^{\parallel}\|_F=|c_t|\|U_t\|_F,
    &\|U_t^{\perp}\|_F=\|U_t\|_F\sqrt{1-c_t^2}.
    \end{aligned}
\end{equation}

The tangential component directly changes the parameter direction, whereas a purely radial component produces no angular displacement by itself. When $c_t<0$, $U_t^{\parallel}$ points opposite to $W_t$; because the update rule applies $-\eta_t U_t$, this component tends to increase the parameter norm. Conversely, when $c_t>0$, the radial component tends to decrease the parameter norm. Even though a radial update does not directly rotate the parameter, it can still affect future optimization indirectly by changing $\|W_t\|_F$, the gradient scale, and the optimizer state. The analysis below therefore focuses on the radial component's direct contribution to the \emph{current one-step angular update}; it does not rule out longer-term indirect effects.

\subsection{Local effect of radial updates on the angular effective learning rate}

By Eq.~\eqref{eq:exact-angular-update}, the angular displacement at the current step satisfies
\begin{equation}
    \tan(\Delta\phi_t)
    =
    \frac{
        \eta_t\sqrt{1-c_t^2}\|U_t\|_F
    }{
        \alpha_t\|W_t\|_F-\eta_tc_t\|U_t\|_F
    }.
    \label{eq:radial-angular-update}
\end{equation}
This expression separates two effects of the radial component. First, $c_t$ appears in the numerator: as $|c_t|$ increases, the tangential magnitude $\|U_t\|_F\sqrt{1-c_t^2}$ decreases, leaving less of the update available to rotate the parameter. Second, $c_t$ appears in the denominator: when $c_t<0$, the denominator increases and the angular displacement is reduced further. Thus, a negative radial component suppresses the one-step angular motion, although the magnitude of this effect depends on the relative scales of $\eta_t$, $\|W_t\|_F$, and $\|U_t\|_F$. To analyze how these factors jointly evolve during training, treat the step index as a continuous variable and differentiate $\log \eta^{\phi}_{\mathrm{eff},t}$. This gives
\begin{equation}
    \frac{\mathrm{d}\log \eta^{\phi}_{\mathrm{eff},t}}{\mathrm{d}t}
    =
    \frac{\dot{\eta}_t}{\eta_t}
    -
    \frac{c_t\dot{c}_t}{1-c_t^2} 
    -
    \frac{
        \dot{\alpha}_t\|W_t\|_F
        +\alpha_t\frac{\mathrm{d}\|W_t\|_F}{\mathrm{d}t}
        -\dot{\eta}_t\|U_t\|_Fc_t
        -\eta_tc_t\frac{\mathrm{d}\|U_t\|_F}{\mathrm{d}t}
        -\eta_t\|U_t\|_F\dot{c}_t
    }{
        \alpha_t\|W_t\|_F-\eta_t\|U_t\|_Fc_t
    }.
    \label{eq:effective-lr-dynamics}
\end{equation}
Equation~\eqref{eq:effective-lr-dynamics} decomposes the evolution of the angular effective learning rate into four sources: the external learning-rate schedule, changes in the parameter--update angle, changes in the parameter norm, and changes in the update norm.

\begin{figure}[htbp]
\centering
\begin{subfigure}[t]{0.48\linewidth}
    \centering
    \includegraphics[width=\linewidth]{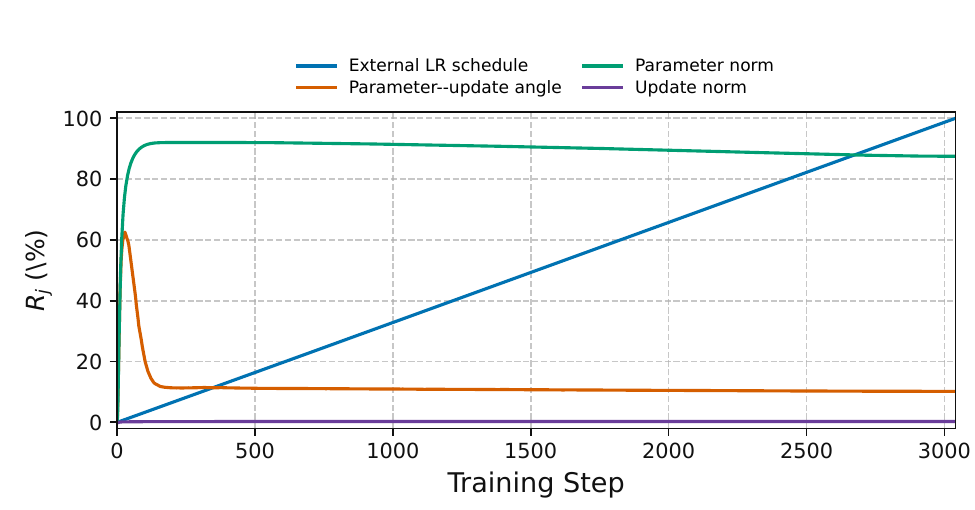}
    \caption{}
    \label{fig:contribution-muonwd}
\end{subfigure}
\hfill
\begin{subfigure}[t]{0.48\linewidth}
    \centering
    \includegraphics[width=\linewidth]{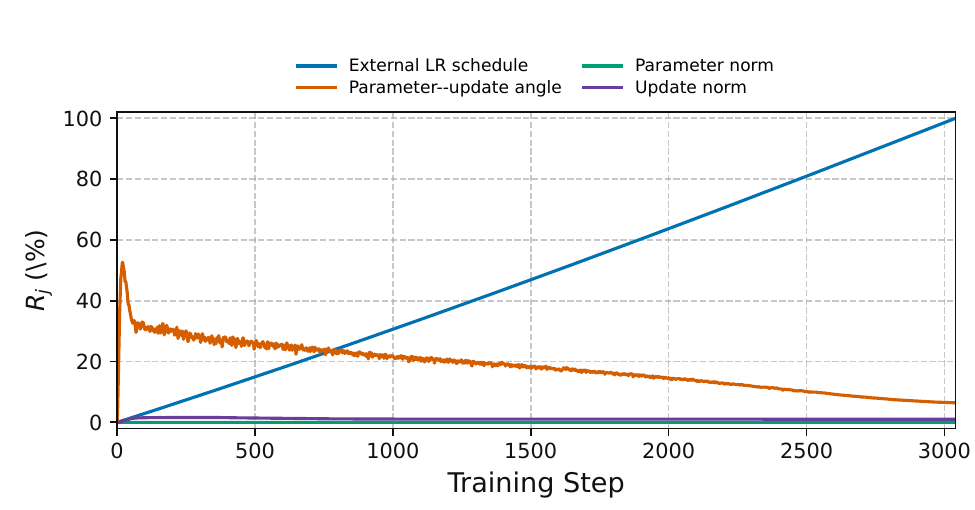}
    \caption{}
    \label{fig:contribution-muonh}
\end{subfigure}
\caption{The decline in learning rate caused by different factors. (a) MuonWD. (b) MuonH.}
\label{FIG1}
\end{figure}

As shown in Fig.~\ref{FIG1}, we further use Eq.~\eqref{eq:effective-lr-dynamics} to quantify the contribution of each factor to the evolution of the angular effective learning rate, using \(R_j=100\left(
1-\frac{\eta^{(j)}(t)}{\eta^{(j-1)}(t)}
\right)\% \) as the metric. Under the training configurations considered here, $c_t$ changes substantially at the beginning of training and then enters a slowly varying negative regime. During this later regime, the terms involving $c_t$ make only a limited instantaneous contribution to Eq.~\eqref{eq:effective-lr-dynamics}; the overall trend is governed primarily by the external learning-rate schedule and the evolution of the parameter norm.

These results do not support the hypothesis that MuonH's two-stage training dynamics are driven primarily by the removal of radial updates. More precisely, the radial component affects the one-step angular displacement and may influence long-term training indirectly through the parameter norm. Within the scope of our experiments, however, its direct effect is not the dominant explanation for the phase-dependent dynamics. The next section examines an alternative mechanism: Hyperball may act chiefly by fixing the scales of the parameter and update, thereby inducing a state-dependent effective-learning-rate schedule.

\section{No Free Lunch: Hyperball as an Implicit Learning-Rate Scheduler}
Having ruled out the direct removal of radial updates as the primary explanation, we interpret Hyperball's characteristic slow-then-fast behavior as a consequence of its effective step-size dynamics. \citet{River} report a related valley-landscape phenomenon when comparing AdamW with Adam: even under the same base learning-rate schedule, AdamW can begin with a higher loss yet ultimately converge to a strictly lower loss. This observation suggests an alternative explanation for Hyperball's behavior, analogous to the effect of weight decay: \textbf{a larger effective step size can delay convergence while improving the solution reached later in training}.

To make this connection explicit, we decompose the Hyperball update into a state-dependent radial rescaling and an update-normalization term:
\begin{equation}
    W_{t+1}=\frac{\|W_t\|_F}{\left\|W_t-\eta u_t \frac{\|W_t\|_F}{\|u_t\|_F}\right\|_F}W_t
    - \frac{\eta\|W_t\|_F}{\left\|W_t-\eta u_t \frac{\|W_t\|_F}{\|u_t\|_F}\right\|_F}
    u_t \frac{\|W_t\|_F}{\|u_t\|_F}.
\end{equation}
Define the state-dependent projection coefficient
\begin{equation}
    \alpha_t^H
    \triangleq
    \frac{\|W_t\|_F}
    {\left\|W_t-\eta u_t \frac{\|W_t\|_F}{\|u_t\|_F}\right\|_F}.
\end{equation}
The update then simplifies to
\begin{equation}
    W_{t+1}=\alpha_t^H W_t- \eta\alpha_t^H u_t \frac{\|W_t\|_F}{\|u_t\|_F}.
\end{equation}
In this form, Hyperball combines a state-dependent radial coefficient with update normalization; the coefficient is determined by the current state rather than by an independently specified weight-decay hyperparameter. Its two-stage behavior may therefore arise from the same step-size mechanism commonly associated with weight decay.

This interpretation is a working hypothesis and requires both theoretical and empirical validation. Weight decay is often understood to increase the angular effective learning rate by controlling parameter-norm growth. Section~\ref{sec:radial-update} indicates that the radial component's direct contribution to the current angular step is small in our setting, suggesting that scale control is the more relevant mechanism. Hyperball fixes the parameter norm and normalizes the update; once the transient variation in $c_t$ subsides, its angular effective learning rate closely tracks the prescribed base learning-rate schedule instead of being attenuated by parameter-norm growth~\citep{MuonH,blake2024umup,kosson2025weightdecay}. Hyperball can thus be viewed as an optimizer with an approximately controlled angular step size, which can produce the same delayed-convergence phenomenon associated with a larger effective learning rate.

To test whether the slower initial convergence of MuonH is attributable to its larger effective step size, we conduct three experiments:
\begin{itemize}
    \item[1.] We visualize and compare the effective-learning-rate trajectories of MuonWD and MuonH.
    \item[2.] We test whether the validation-loss dynamics of Hyperball and non-Hyperball optimizers can be transformed into one another by changing only the learning-rate schedule.
    \item[3.] In language-model pretraining experiments, we evaluate whether alternative learning-rate schedules can accelerate the convergence of MuonH.
\end{itemize}

\subsection{Empirical evolution of the effective learning rate under different schedules}
The effective learning rates of non-Hyperball optimizers such as AdamW and MuonWD have been characterized by norm-based analyses. Section~\ref{sec:radial-update} extends this picture to the non-orthogonal regime $c_t<0$. Here we directly compare MuonWD and MuonH. The purpose is to establish two empirical facts: MuonH operates at a larger angular effective learning rate in the configurations considered, and its effective learning rate remains approximately proportional to the externally specified learning rate.

\begin{figure}[htbp]
\centering
\begin{subfigure}[t]{0.24\linewidth}
    \centering
    \includegraphics[width=\linewidth]{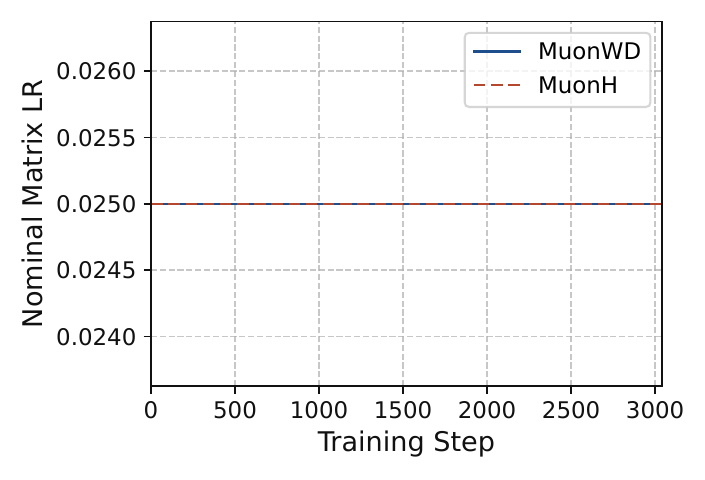}
    \caption{}
    \label{fig:ex1-fixed-lr}
\end{subfigure}
\begin{subfigure}[t]{0.24\linewidth}
    \centering
    \includegraphics[width=\linewidth]{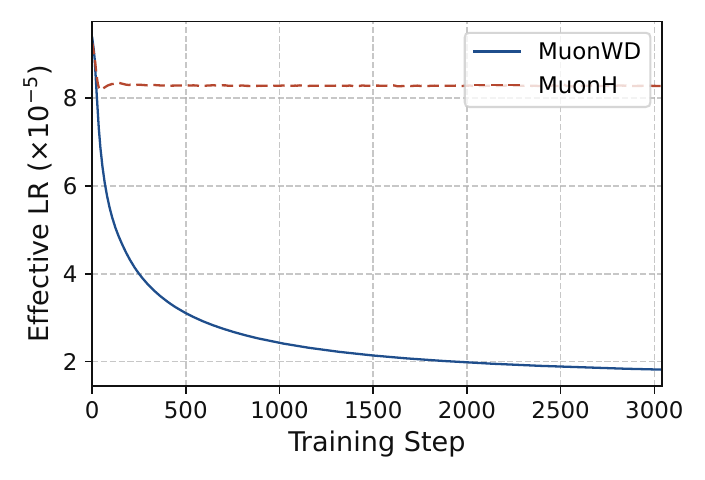}
    \caption{}
    \label{fig:ex1-fixed-effective-lr}
\end{subfigure}
\begin{subfigure}[t]{0.24\linewidth}
    \centering
    \includegraphics[width=\linewidth]{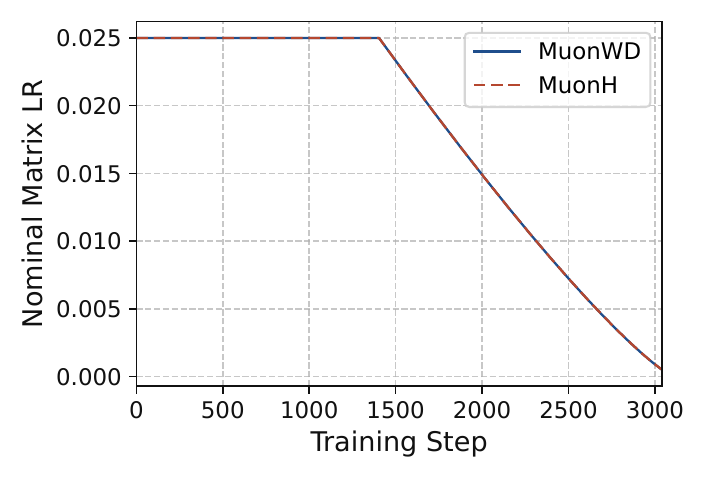}
    \caption{}
    \label{fig:ex1-decay-lr}
\end{subfigure}
\begin{subfigure}[t]{0.24\linewidth}
    \centering
    \includegraphics[width=\linewidth]{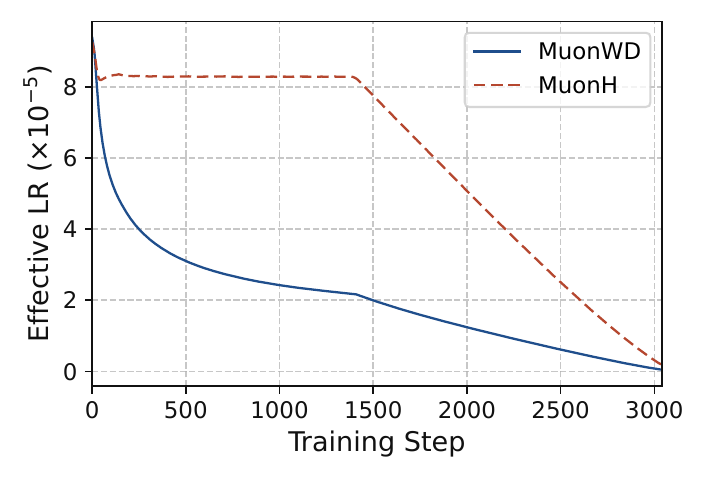}
    \caption{}
    \label{fig:ex1-decay-effective-lr}
\end{subfigure}
\caption{Learning-rate schedules and the resulting effective learning rates. (a) Constant learning-rate schedule. (b) Effective-learning-rate comparison under the constant schedule. (c) Constant-then-decay schedule. (d) Effective-learning-rate comparison under the constant-then-decay schedule.}
\label{fig:learning-aqa}
\end{figure}

Figure~\ref{fig:learning-aqa} shows the learning-rate schedules and their corresponding effective-learning-rate trajectories. Under a constant base learning rate, MuonWD's effective learning rate decays rapidly as its parameter norm grows. MuonH, in contrast, maintains an approximately constant effective learning rate after an initial transient caused by the abrupt evolution of $c_t$. Under a linearly decaying base learning rate, MuonH's effective learning rate remains approximately proportional to that schedule.

\subsection{Transforming the dynamics of Hyperball and non-Hyperball optimizers}
\begin{figure}[htbp]
\centering
\begin{subfigure}[t]{0.24\linewidth}
    \centering
    \includegraphics[width=\linewidth]{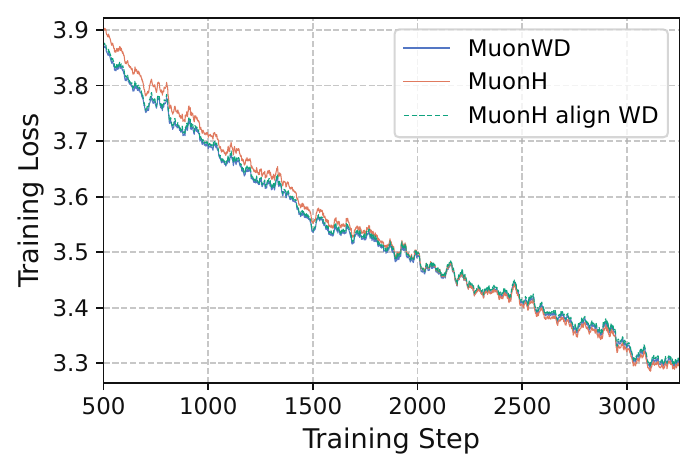}
    \caption{}
    \label{fig:ex2-hyperball-to-weight-decay}
\end{subfigure}
\begin{subfigure}[t]{0.24\linewidth}
    \centering
    \includegraphics[width=\linewidth]{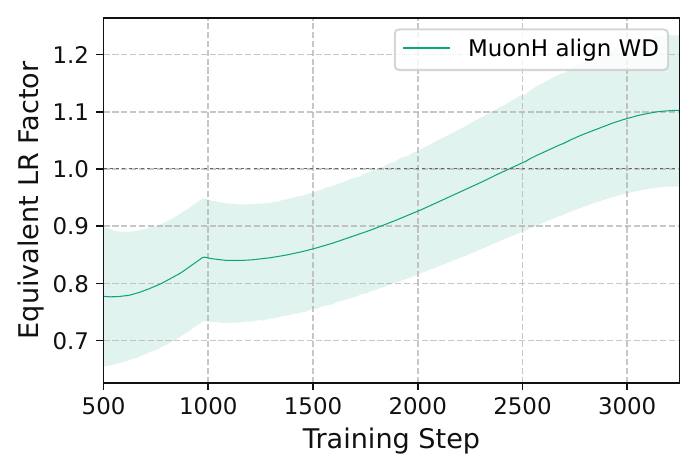}
    \caption{}
    \label{fig:ex2-hyperball-lr-factor}
\end{subfigure}
\begin{subfigure}[t]{0.24\linewidth}
    \centering
    \includegraphics[width=\linewidth]{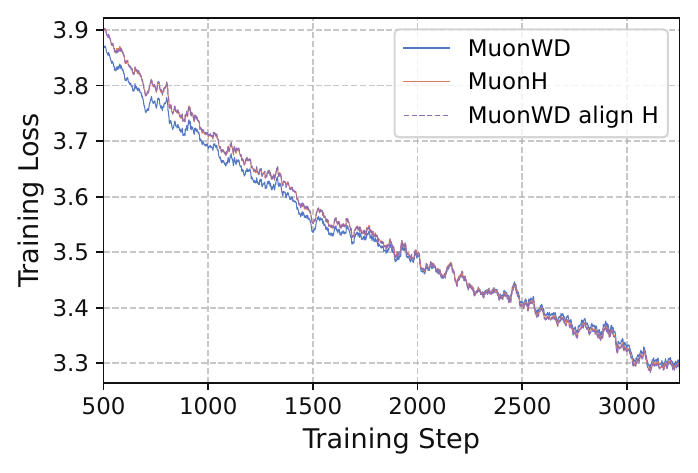}
    \caption{}
    \label{fig:ex2-weight-decay-to-hyperball}
\end{subfigure}
\begin{subfigure}[t]{0.24\linewidth}
    \centering
    \includegraphics[width=\linewidth]{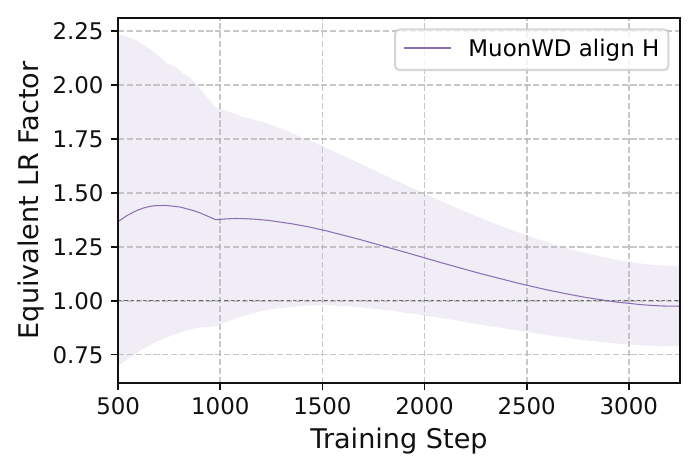}
    \caption{}
    \label{fig:ex2-weight-decay-lr-factor}
\end{subfigure}
\caption{Mutual alignment of Hyperball and non-Hyperball dynamics. (a) Validation loss when MuonH is aligned with MuonWD. (b) MuonH learning-rate alignment factors during this experiment. (c) Validation loss when MuonWD is aligned with MuonH. (d) MuonWD learning-rate alignment factors during the reverse experiment. Shaded regions in (b) and (d) show variation across layers.}
\label{fig:learning-444}
\end{figure}

We next test whether the effective learning rate is the principal difference between Hyperball and non-Hyperball optimizers. Because fixed parameter and update norms can change multiple aspects of a high-dimensional optimization trajectory, observational comparisons alone cannot isolate their contribution. We therefore perform a controlled ablation in which the optimizer is held fixed and only its learning rate is changed so as to match the target angular effective learning rate at every step. Denote the effective learning rates of the non-Hyperball and Hyperball optimizers under the reference schedule by $\eta_{\mathrm{eff},t}^{\phi}$ and $\eta_{\mathrm{eff},t}^{H,\phi}$, respectively. To align MuonWD with MuonH, we proceed as follows:
\begin{itemize}
    \item[1.] Run MuonH and record its angular effective learning rate $\eta_{\mathrm{eff},t}^{H,\phi}$ at every step.
    \item[2.] Before each MuonWD update, record $c_t$, $W_t$, and $U_t$, and solve
    \begin{equation}
        \eta_{\mathrm{eff},t}^{H,\phi}
        \approx
        \frac{\eta_t\sqrt{1-c_t^2}}
        {\alpha_t\|W_t\|_F-c_t\eta_t\|U_t\|_F}
    \end{equation}
    for the step-specific learning rate $\eta_t$.
    \item[3.] Use the resulting $\eta_t$ for the current MuonWD update.
\end{itemize}
We apply the analogous procedure in the reverse direction to align MuonH with MuonWD. The results in Fig.~\ref{fig:learning-444} show that the validation-loss trajectories of the two optimizer families can be transformed into one another to a substantial degree by learning-rate alignment alone. This finding supports the interpretation that, in the settings studied, Hyperball's dominant effect is an implicit state-dependent learning-rate schedule rather than an intrinsically superior update direction.
\subsection{Accelerating Hyperball-Style Optimizers}

Building on the observations validated above, we accelerate the MuonH optimizer through learning-rate scheduling. Following the river-valley landscape interpretation of \citet{River}, \textbf{the initially large step size delays convergence until the later stages of training}. Moreover, the effective learning rate of Hyperball-style optimizers is inherently larger than that of conventional optimizers during the early stage. We therefore adopt more aggressive learning-rate decay schedules to accelerate MuonH. We consider the following schedules:
\begin{itemize}
    \item[1.] The linear learning-rate decay schedule, under which $\eta_t$ decreases linearly with $t$:
    \begin{equation}
        \eta_{\mathrm{lin}}(t)=
        \eta_0\left(1-\dfrac{t}{T}\right),
        \quad 0\le t\le T.
    \end{equation}

    \item[2.] The exponential learning-rate decay schedule, under which $\eta_t$ decreases exponentially with $t$:
    \begin{equation}
        \eta_{\mathrm{exp}}(t)=
        \eta_0\,
        \dfrac{\exp\left(-\lambda \dfrac{t}{T}\right)-\exp(-\lambda)}
        {1-\exp(-\lambda)},
        \quad 0\le t\le T,\ \lambda>0.
    \end{equation}

    \item[3.] The polynomial learning-rate decay schedule, under which $\eta_t$ decreases polynomially with $t$:
    \begin{equation}
        \eta_{\mathrm{pow}}(t)=
        \eta_0\left[1-\left(\dfrac{t}{T}\right)^p\right],
        \quad 0\le t\le T,\ p>0.
    \end{equation}
\end{itemize}

In the comparisons below, \emph{MuonH-Z} denotes the minus-square-root learning-rate schedule proposed by Zhanpeng Zhou in \href{https://github.com/KellerJordan/modded-nanogpt/pull/343}{PR~\#343}, which reaches the target validation loss of $3.28$ in 3175 steps. \emph{MuonH-Ours} denotes our power-$0.4$ schedule released in \href{https://github.com/KellerJordan/modded-nanogpt/pull/345}{PR~\#345}, which reaches the same target in 3150 steps.

\begin{figure}[htbp]
\centering
\begin{subfigure}[t]{0.24\linewidth}
    \centering
    \includegraphics[width=\linewidth]{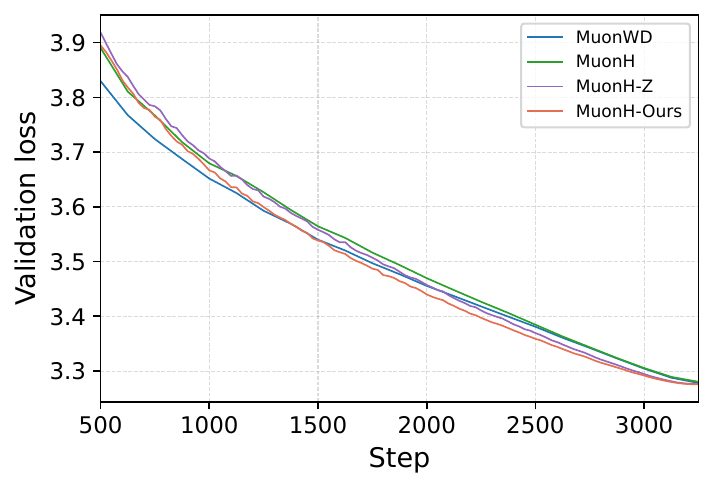}
    \caption{}
    \label{fig:pretraining-validation-comparison}
\end{subfigure}
\begin{subfigure}[t]{0.24\linewidth}
    \centering
    \includegraphics[width=\linewidth]{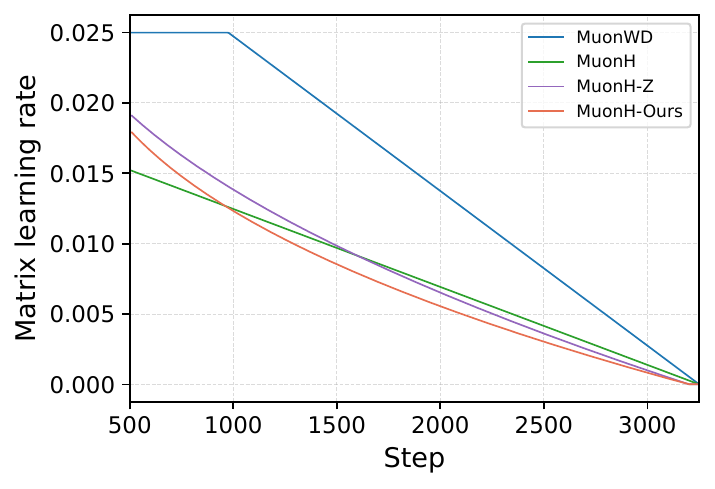}
    \caption{}
    \label{fig:pretraining-lr-schedules}
\end{subfigure}
\begin{subfigure}[t]{0.24\linewidth}
    \centering
    \includegraphics[width=\linewidth]{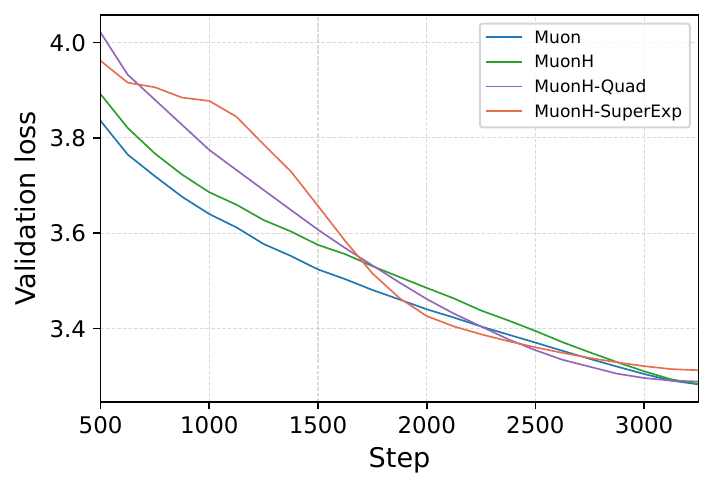}
    \caption{}
    \label{fig:pretraining-late-validation}
\end{subfigure}
\begin{subfigure}[t]{0.24\linewidth}
    \centering
    \includegraphics[width=\linewidth]{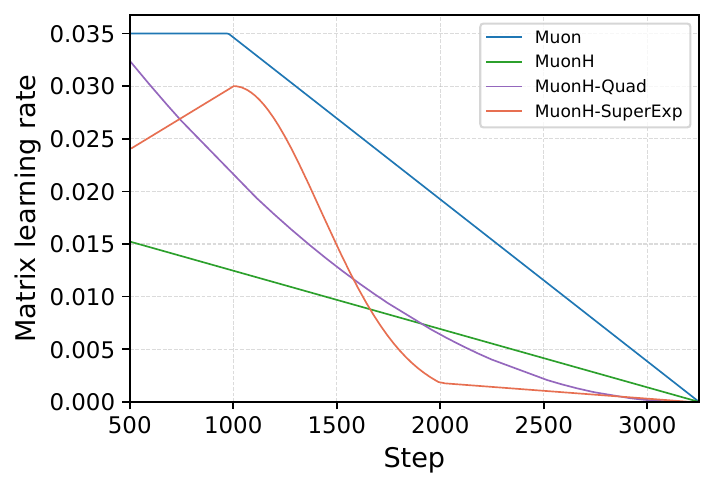}
    \caption{}
    \label{fig:fig1-4}
\end{subfigure}
\caption{Validation loss and learning-rate schedules under different decay strategies. (a) Validation losses of three MuonH variants in the pretraining experiments. (b) The corresponding learning-rate schedules for (a). (c) Validation losses under an additional set of learning-rate schedules. (d) The corresponding learning-rate schedules for (c). Even when MuonH surpasses Muon earlier in training, Muon can overtake it later.}
\label{fig:learning-4044}
\end{figure}

Because MuonH maintains a constant angular velocity, we adopt the polynomial learning-rate decay schedule, which provides a more aggressive decay, in our language-model pretraining experiments. As shown in Fig.~\ref{fig:learning-4044}, our method reaches a validation loss below $3.28$ faster than the compared baselines. However, our search experiments also reveal a new observation: making MuonH faster during the early stage does not necessarily guarantee better convergence in the later stage. When a more aggressive learning-rate decay schedule is used to induce earlier convergence, MuonH performs substantially worse during the later stage than standard MuonWD and MuonH with other learning-rate schedules. Thus far, we have found no direct evidence that MuonH consistently outperforms MuonWD throughout the entire optimization process.

This observation provides a more fundamental insight into MuonH: effectively using MuonH requires a new learning-rate scheduling strategy. Directly conducting scaling experiments with MuonH may even degrade model performance. Maintaining a constant angular velocity does not directly reduce the complexity of the optimization problem; instead, it makes learning-rate scheduling even more important in the optimization dynamics of MuonH.

\section{Discussion}

\subsection{Limitations}

Our study has several limitations. First, the theoretical analysis is limited to a numerical sensitivity analysis of the effective learning rate and an examination of the conventional effective-learning-rate estimate. The river-valley landscape provides an interpretation of the slow-then-fast convergence dynamics of Hyperball-style optimizers~\citep{River,MuonH}, but a more complete theoretical characterization remains open. Second, our current experiments focus on dense language-model pretraining. Extending the analysis to mixture-of-experts (MoE) architectures will be important for determining whether the observed dynamics persist under sparse activation and expert routing. Third, the optimizer comparison is currently restricted to Muon, MuonWD, and MuonH. Future experiments will broaden this comparison to other optimizer families, including AdamW and its Hyperball counterpart AdamH. Finally, our evaluation primarily considers pretraining validation loss and optimization dynamics. A more comprehensive assessment should include a wider range of downstream tasks and metrics to determine whether the observed optimization behavior translates into improvements in generalization and practical model quality.

Our analysis also relies on a scale-invariance assumption that does not strictly hold in practical training. Interestingly, the two properties implied by scale invariance---that the stochastic gradient is orthogonal to the model parameters and that the gradient magnitude decreases as the model norm increases---are nevertheless observed in practice~\citep{salimans2016weight,ballsun}. Understanding when this approximation is sufficiently accurate, and when its violations materially affect the predicted dynamics, requires further investigation.

\subsection{Future Directions}

Beyond these limitations, several broader questions emerge from our investigation.

\textbf{How should the learning rate be scheduled?}
Learning-rate scheduling remains a largely open problem. Recent studies have primarily investigated approaches such as WSD~\citep{River,li2020reconciling}, schedule-free methods~\citep{schedu,schedu++}, and the derivation of learning-rate schedules from functional rates~\citep{Leiwu1,leiwu2}. From the current perspective, a Hyperball-style optimizer incorporates a heuristic learning-rate schedule based on weight constraints. Although this mechanism does not reduce the complexity of the scheduling problem, it does not imply that the contribution of Hyperball-style optimizers is itself merely incremental. Their specific update dynamics may provide deeper insights into learning-rate scheduling.

Notably, the effective learning rate of a Hyperball-style optimizer is proportional to its scheduled learning rate. Therefore, using Hyperball-style optimizers as a testbed for studying learning-rate schedules can largely avoid the confounding effects introduced by the coupling between the learning rate and weight decay. This property makes them a particularly promising experimental foundation for investigating learning-rate scheduling.

\textbf{Does a two-stage learning-rate schedule exist?}
For non-Hyperball-style optimizers, the effective learning rate decreases rapidly during the early stage as the model norm grows. During the middle and later stages of training, steady-state analysis yields
$\eta_{\mathrm{eff}}^\phi\propto\sqrt{\eta\lambda}$~\citep{su2024spectral,su2025adamupdate,su2025adamwrms1,su2025adamwrms2}.
By contrast, the effective learning rate of a Hyperball-style optimizer satisfies
$\eta_{\mathrm{eff}}^{H,\phi}\propto\eta$.
It remains to be investigated whether, during the later stages of optimization, the effective learning rate of a Hyperball-style optimizer can become smaller than that of a conventional optimizer and thereby hinder its performance.

\textbf{Rethinking first-order optimization.}
Beyond optimizers based on geometric constraints, another promising direction is to use first-order information to approximate second-order Hessian information or even third-order information for improved optimization. Existing approaches can be broadly divided into two categories: methods that use gradient information of different orders to estimate the step size~\citep{ada,optmuon}, and methods that use such information to estimate the update direction~\citep{new1,new2,huanran,SSO}. In the context of learning-rate scheduling, it may be necessary to investigate more deeply how gradient or even Hessian information can be used to characterize the coupling between the step size and update direction.

\section{Conclusion}

Starting from the effective learning rate, we first examine its numerical sensitivity to perturbations introduced by radial updates and show that it is only weakly affected by such updates. Based on this observation, we challenge existing heuristic explanations of the dynamics of Hyperball-style optimizers. We then build on the constant-angular-velocity property of Hyperball-style optimizers and progressively show that their characteristic update dynamics fundamentally arise from the larger effective learning rate induced during the early stage of optimization.

To validate this interpretation, we align the effective learning rates of the optimizers and show that the dynamics of a Hyperball-style optimizer can be reproduced simply by modifying the learning-rate schedule of Muon.
\section*{Acknowledgments}
We thank Huaqing Zhang, Tian Xie, Guangyu Chen, Huanran Chen, Ganzhao Yuan, and Zhenjie Zhou for their insightful discussions during the development of this work.

\newpage

\bibliography{hyperball.bib}

@article{ballsun,
  title={Spherical motion dynamics: Learning dynamics of neural network with normalization, weight decay, and sgd},
  author={Wan, Ruosi and Zhu, Zhanxing and Zhang, Xiangyu and Sun, Jian},
  journal={arXiv preprint arXiv:2006.08419},
  year={2020}
}

@article{1,
  title={Power lines: Scaling laws for weight decay and batch size in llm pre-training},
  author={Bergsma, Shane and Dey, Nolan and Gosal, Gurpreet and Gray, Gavia and Soboleva, Daria and Hestness, Joel},
  journal={Advances in Neural Information Processing Systems},
  volume={38},
  pages={125153--125188},
  year={2026}
}

@inproceedings{2,
  title={Group normalization},
  author={Wu, Yuxin and He, Kaiming},
  booktitle={Proceedings of the European conference on computer vision (ECCV)},
  pages={3--19},
  year={2018}
}

@article{MuonH,
  title={Fantastic Pretraining Optimizers and Where to Find Them II: Hyperball Optimization},
  author={Wen, Kaiyue and Dang, Xingyu and Lyu, Kaifeng and Ma, Tengyu and Liang, Percy},
  journal={arXiv preprint arXiv:2606.16899},
  year={2026}
}

@article{rmnp,
  title={Rmnp: Row-momentum normalized preconditioning for scalable matrix-based optimization},
  author={Deng, Shenyang and Ouyang, Zhuoli and Pang, Tianyu and Liu, Zihang and Jin, Ruochen and Yu, Shuhua and Yang, Yaoqing},
  journal={arXiv preprint arXiv:2603.20527},
  year={2026}
}

@article{eff1,
  title={L2 regularization versus batch and weight normalization},
  author={Van Laarhoven, Twan},
  journal={arXiv preprint arXiv:1706.05350},
  year={2017}
}

@article{eff2,
  title={Norm matters: efficient and accurate normalization schemes in deep networks},
  author={Hoffer, Elad and Banner, Ron and Golan, Itay and Soudry, Daniel},
  journal={Advances in Neural Information Processing Systems},
  volume={31},
  year={2018}
}

@article{eff3,
  title={Three mechanisms of weight decay regularization},
  author={Zhang, Guodong and Wang, Chaoqi and Xu, Bowen and Grosse, Roger},
  journal={arXiv preprint arXiv:1810.12281},
  year={2018}
}

@inproceedings{River,
  title={Understanding warmup-stable-decay learning rates: A river valley loss landscape view},
  author={Wen, Kaiyue and Li, Zhiyuan and Wang, Jason and Hall, David and Liang, Percy and Ma, Tengyu},
  booktitle={International Conference on Learning Representations},
  volume={2025},
  pages={42840--42885},
  year={2025}
}

@article{schedu++,
  title={Schedulefree+: Scaling learning-rate-free \& schedule-free learning to large language models},
  author={Defazio, Aaron},
  journal={arXiv preprint arXiv:2605.19095},
  year={2026}
}

@article{schedu,
  title={The road less scheduled},
  author={Defazio, Aaron and Yang, Xingyu and Mehta, Harsh and Mishchenko, Konstantin and Khaled, Ahmed and Cutkosky, Ashok},
  journal={Advances in Neural Information Processing Systems},
  volume={37},
  pages={9974--10007},
  year={2024}
}

@article{Leiwu1,
  title={Optimal Learning-Rate Schedules under Functional Scaling Laws: Power Decay and Warmup-Stable-Decay},
  author={Li, Binghui and Wang, Zilin and Chen, Fengling and Zhao, Shiyang and Zheng, Ruiheng and Wu, Lei},
  journal={arXiv preprint arXiv:2602.06797},
  year={2026}
}

@article{leiwu2,
  title={Fast catch-up, late switching: Optimal batch size scheduling via functional scaling laws},
  author={Wang, Jinbo and Li, Binghui and Zhou, Zhanpeng and Wang, Mingze and Sun, Yuxuan and Zhang, Jiaqi and Cai, Xunliang and Wu, Lei},
  journal={arXiv preprint arXiv:2602.14208},
  year={2026}
}

@article{optmuon,
  title={OptMuon: Closed-Loop Orthogonalized Momentum Methods for Stochastic Optimization with Zero-Noise Optimality},
  author={Yuan, Ganzhao},
  journal={arXiv preprint arXiv:2606.08783},
  year={2026}
}

@misc{ada,
      title={Adafactor: Adaptive Learning Rates with Sublinear Memory Cost}, 
      author={Noam Shazeer and Mitchell Stern},
      year={2018},
      eprint={1804.04235},
      archivePrefix={arXiv},
      primaryClass={cs.LG},
      url={https://arxiv.org/abs/1804.04235}, 
}

@article{huanran,
  title={Nexus: Same Pretraining Loss, Better Downstream Generalization via Common Minima},
  author={Chen, Huanran and Zhang, Huaqing and Li, Xiao and Dong, Yinpeng and Shen, Ke and Zhu, Jun},
  journal={arXiv preprint arXiv:2604.09258},
  year={2026}
}

@article{blake2024umup,
  author  = {Blake, Charlie and Eichenberg, Christian and Dean, Jeff and Balles, Lukas and Prince, L. Y. and Deiseroth, B. and Cruz-Salinas, A. F. and Luschi, Carlo and Weinbach, S. and Orr, David},
  title   = {u-{$\mu$P}: The Unit-Scaled Maximal Update Parametrization},
  journal = {arXiv preprint arXiv:2407.17465},
  year    = {2024},
  url     = {https://arxiv.org/abs/2407.17465}
}

@misc{cesista2025stiefel,
  author = {Cesista, F. L.},
  title  = {Heuristic Solutions for Steepest Descent on the Stiefel Manifold},
  year   = {2025},
  url    = {https://leloykun.github.io/ponder/steepest-descent-stiefel/}
}

@article{kosson2025weightdecay,
  author  = {Kosson, A. and Welborn, J. and Liu, Y. and Jaggi, M. and Chen, X.},
  title   = {Weight Decay May Matter More Than {$\mu$P} for Learning Rate Transfer in Practice},
  journal = {arXiv preprint arXiv:2510.19093},
  year    = {2025},
  url     = {https://arxiv.org/abs/2510.19093}
}

@inproceedings{li2020reconciling,
  author    = {Li, Z. and Lyu, K. and Arora, S.},
  title     = {Reconciling Modern Deep Learning with Traditional Optimization Analyses: The Intrinsic Learning Rate},
  booktitle = {Advances in Neural Information Processing Systems},
  year      = {2020},
  url       = {https://arxiv.org/abs/2010.02916},
  note      = {arXiv:2010.02916}
}

@article{li2025normuon,
  author  = {Li, Z. and Liu, L. and Liang, C. and Chen, W. and Zhao, T.},
  title   = {{NorMuon}: Making {Muon} More Efficient and Scalable},
  journal = {arXiv preprint arXiv:2510.05491},
  year    = {2025},
  url     = {https://arxiv.org/abs/2510.05491}
}

@inproceedings{salimans2016weight,
  author    = {Salimans, Tim and Kingma, Diederik P.},
  title     = {Weight Normalization: A Simple Reparameterization to Accelerate Training of Deep Neural Networks},
  booktitle = {Advances in Neural Information Processing Systems},
  year      = {2016},
  url       = {https://arxiv.org/abs/1602.07868}
}

@misc{su2024spectral,
  author = {Su, J.},
  title  = {Thinking about Spectral Norm Gradient and Spectral Weight Decay},
  year   = {2024},
  url    = {https://kexue.fm/archives/10648}
}

@misc{su2025adamupdate,
  author = {Su, J.},
  title  = {Why {Adam}'s Update {RMS} Is 0.2?},
  year   = {2025},
  url    = {https://kexue.fm/archives/11267}
}

@misc{su2025adamwrms1,
  author = {Su, J.},
  title  = {{AdamW} Weight {RMS} Asymptotics (Part {I})},
  year   = {2025},
  url    = {https://kexue.fm/archives/11307}
}

@misc{su2025adamwrms2,
  author = {Su, J.},
  title  = {{AdamW} Weight {RMS} Asymptotics (Part {II})},
  year   = {2025},
  url    = {https://kexue.fm/archives/11404}
}

@article{new1,
  title={Understanding optimization in deep learning with central flows},
  author={Cohen, Jeremy M and Damian, Alex and Talwalkar, Ameet and Kolter, J Zico and Lee, Jason D},
  journal={arXiv preprint arXiv:2410.24206},
  year={2024}
}

@article{new2,
  title={Label noise sgd provably prefers flat global minimizers},
  author={Damian, Alex and Ma, Tengyu and Lee, Jason D},
  journal={Advances in Neural Information Processing Systems},
  volume={34},
  pages={27449--27461},
  year={2021}
}

@article{SSO,
  title={Controlled llm training on spectral sphere},
  author={Xie, Tian and Luo, Haoming and Tang, Haoyu and Hu, Yiwen and Liu, Jason Klein and Ren, Qingnan and Wang, Yang and Zhao, Wayne Xin and Yan, Rui and Su, Bing and others},
  journal={arXiv preprint arXiv:2601.08393},
  year={2026}
}

\newpage
\appendix

\end{CJK*}
\end{document}